\documentclass{article}

\usepackage[final]{neurips_2019_ml4ps}

\usepackage[utf8]{inputenc} %
\usepackage[T1]{fontenc}    %
\usepackage{hyperref}       %
\usepackage{url}            %
\usepackage{booktabs}       %
\usepackage{amsfonts}       %
\usepackage{amsmath}
\usepackage{nicefrac}       %
\usepackage{microtype}      %
\usepackage{color}
\usepackage{graphicx}
\usepackage[autolanguage]{numprint}

\setcitestyle{numbers}
\setcitestyle{square}
\setcitestyle{comma}
\newcommand{\bx}{\mathbf{x}}
\newcommand{\bv}{\mathbf{v}}
\newcommand{\bm}{\mathbf{m}}
\newcommand{\bb}{\mathbf{b}}
\newcommand{\bc}{\mathbf{c}}
\newcommand{\bd}{\mathbf{d}}
\newcommand{\bs}{\mathbf{s}}

\title{Scalable Extreme Deconvolution}

\author{
  James A. Ritchie\\
  School of Informatics\\
  University of Edinburgh\\
   \texttt{james.ritchie@ed.ac.uk} \\
  \And
  Iain Murray\\
  School of Informatics\\
  University of Edinburgh\\
   \texttt{i.murray@ed.ac.uk} \\
}

\begin{document}

\maketitle

\begin{abstract}

The Extreme Deconvolution method fits a probability density to a dataset where each observation has Gaussian noise added with a known sample-specific covariance, originally intended for use with astronomical datasets.
The existing fitting method is batch EM, which would not normally be applied to large datasets such as the Gaia catalog containing noisy observations of a billion stars.
We propose two minibatch variants of extreme deconvolution, based on an online variation of the EM algorithm, and direct gradient-based optimisation of the log-likelihood, both of which can run on GPUs.
We demonstrate that these methods provide faster fitting, whilst being able to scale to much larger models for use with larger datasets.

\end{abstract}

\section{Introduction}

Extreme deconvolution is a method that fits Gaussian Mixture Models (GMMs) to noisy data where we know the covariance of the Gaussian noise added to each sample~\cite{bovyExtremeDeconvolutionInferring2011}.
The method was originally developed to perform probabilistic density estimation on the dataset of stellar velocities produced by the Hipparcos satellite~\cite{perrymanHipparcosCatalogue1997}.
The Hipparcos catalogue consists of astrometric solutions (positions and velocities on the sky) and photometry (light intensity) for 118,218 stars, with associated noise covariances provided for each entry.

The successor to the Hipparcos mission, Gaia, aims to produce an even larger catalogue, with entries for an estimated 1 billion astronomical objects~\cite{collaborationGaiaMission2016}.
Previous work using an extreme deconvolution model on the Gaia catalogue %
worked with a subset of the data and restricted the number of mixture components,
but the intention is to fit models with the full dataset
\cite{andersonImprovingGaiaParallax2018}.
The existing extreme deconvolution algorithm makes a full pass over the dataset before it can update parameters, and the reference implementation requires all the data to fit in memory. To fit such large datasets in reasonable time, we would normally use stochastic or online methods, with updates based on minibatches of data to make the methods practical on GPUs \cite{bottou2018}.

In this work, we develop two minibatch methods for fitting the extreme deconvolution model
based on 1) an online variation of the Expectation-Maximisation (EM) algorithm, and 2) a gradient optimizer.
Our implementations can run on GPUs, and provide comparable density estimates to the existing method, whilst being much faster to train.

\section{Background}

The aim of extreme deconvolution is to perform density estimation on a noisy $d$-dimensional dataset $\{\bx_i\}_{i=0}^N$, where $\bx_i$ was generated by adding zero-mean Gaussian noise $\epsilon_i$ with known per-datapoint covariance $S_i$ to a projection $R_i$ of a true value $\bv_i$,
\begin{equation}
  \bx_i = R_i\bv_i + \epsilon_i,\quad  \epsilon_i \sim \mathcal{N}(\mathbf{0}, S_i).
\end{equation}
We assume that $\bv_i$ can be modelled by a mixture of Gaussians with $K$ components,
\begin{equation}
p(\bv_i \mid \theta) = \sum_j^K \alpha_j \,\mathcal{N}(\bv \mid \bm_j, V_j), \quad \theta = \{\alpha_j, \bm_j, V_j\}_{j=1} ^ K,
\end{equation}
parameterised by mixture weight $\alpha_j$, mean $\bm_j$ and covariance $V_j$.
As the noise model is Gaussian and the model of the underlying density is a mixture of Gaussians, the probability of $\bx_i$ is also a Gaussian mixture.
The total log-likehood of the model is
\begin{equation}
\mathcal{L}(\theta) = \sum_i^N \log \sum_j^K \alpha_j\,\mathcal{N}(\bx_i \mid R_i\bm_j, T_{ij}), \quad T_{ij} = R_iV_jR_i^T + S_i.
\end{equation}

Missing data can be handled either by making $R_i$ rank-deficient, or by setting elements of the covariance matrix $S_i$ to very large values.

\section{Methods}

\subsection{Minibatch Expectation-Maximisation}
\label{sec:minibatch-em}
The original method of fitting the extreme deconvolution model used a modification of the Expectation-Maximisation (EM) algorithm for mixture models~\cite{dempsterMaximumLikelihoodIncomplete1977}.
Here we describe a minibatch version of this algorithm based on~\citet{cappeOnlineExpectationMaximization2009}'s online EM algorithm for latent data models.
At each iteration $t$, we compute the sufficient statistics of the latent data $\bv_i$ for each component $j$ in the minibatch of size $M$, using our current estimate of the parameters,
\begin{equation}
r_{ij} = \frac{\alpha_{j} \,\mathcal{N}(\bx_i \mid R_i\bm_j, T_{ij})}{\sum_k \alpha_k \,\mathcal{N}(\bx_i \mid R_i\bm_k, T_{ik})}, \ 
\bb_{ij} = m_j + V_j R_i^T T_{ij}^{-1}(\bx_i - R_i \bm_j), \ 
B_{ij} = V_j - V_j R_i^T T_{ij}^{-1}R_iV_j.
\end{equation}
The $r_{ij}$ term is the posterior probability of datapoint $\bx_i$ coming from component $j$.
The $\bb_{ij}$ and $B_{ij}$ terms result from the fact that $\bx_i$ and $\bv_i$ are jointly Gaussian, so the distribution of $\bv_i$ conditioned on $\bx_i$ is also Gaussian with mean $\bb_{ij}$ and covariance $B_{ij}$.
The expected sufficient statistics are then summed together over the minibatch,
\begin{equation}
q_{jt} = \sum_i r_{ijt}, \quad
\bs_{jt} = \sum_i r_{ijt} \bb_{ijt}, \quad
S_{jt} = \sum_i r_{ijt} [\bb_{ijt}\bb_{ijt}^T + B_{ijt}].
\end{equation}
Stochastic estimates $\hat{\phi}_{jt} $ of the sums of sufficient statistics over the whole dataset are then updated with a sufficiently small step size $\lambda$,
\begin{equation}
\hat{\phi}_{jt} = (1 - \lambda)\hat{\phi}_{j(t-1)} + \lambda \phi_{jt},\quad
\phi_{jt} = \{q_{jt}, \bs_{jt}, S_{jt} \},\quad
\hat{\phi}_{jt} = \{\hat{q}_{jt}, \hat{\bs}_{jt}, \hat{S}_{jt} \}. \label{eq:sums}
\end{equation}
Finally, we normalise the updated sums of expected sufficient statistics to get updated estimates of the parameters,
\begin{equation}
\alpha_{j} = \frac{\hat{q}_{jt}}{M}, \quad
\bm_{j} = \frac{\hat{\bs}_{jt}}{\hat{q}_{jt}}, \quad
V_{j} = \frac{\hat{S}_{jt}}{\hat{q}_{jt}} - \bm_{j} \bm_{j}^T.
\label{eqn:mstep}
\end{equation}
This procedure is repeated with new randomly-ordered minibatches until convergence.
If we set $\lambda=1$ and replace each minibatch with the entire dataset,
then the update corresponds to the original batch fitting method.
Numerically however, the update for $V_j$, as written in \eqref{eqn:mstep}, is inadvisable compared to the batch update given in \cite{bovyExtremeDeconvolutionInferring2011}. There is likely to be catastrophic cancellation if the variances of the components are small relative to the means, especially if single precision floats are used, as is standard with GPU computation.
In Appendix~\ref{apx:variance-rewrite} we show how the minibatch version of this update can be rewritten in a
more numerically stable form.

\subsection{Stochastic Gradient Descent}

An alternative to EM-based methods is to optimise the log-likelihood directly.
The optimization is constrained, because the mixture weights $a_j$ are positive and sum to $1$, and the covariances $V_j$ are positive definite.
Directly fitting the log-likelihood with unconstrained gradient-based optimisers requires a transformation of the parameters to remove the constraints~\cite{williams1996}.
The mixture weights $\alpha_j$ can be parameterised by taking the softmax of an unconstrained vector $\mathbf{z}$, and the covariances $V_j$ represented by its lower triangular Cholesky decomposition $L_j$, where the diagonal elements $qq$ of $L_j$ are constrained positive by taking the exponential of unconstrained elements $\tilde{L}_q$,
\begin{equation}
\alpha_j = \frac{e^{z_j}}{\sum_{k=1}^K e^{z_k}}, \quad
V_j = L_jL_j^T, \quad
(L_j)_{qq} = \exp({\tilde{L}_q}).
\end{equation}
Having removed the constraints, we can optimise the likelihood using any standard minibatch gradient-based optimiser.

For a standard Gaussian mixture model, gradient based optimization has a scaling advantage over EM\@. There is no need to form the $D\!\times\!D$ covariance matrix $V_j$, since the Gaussian density can be evaluated directly from the Cholesky factor $L_j$ in $O(D^2)$, whereas an EM update is $O(D^3)$.
Unfortunately SGD updates are also $O(D^3)$ for the extreme deconvolution model, as we need to form the covariance $T_{ij}$ for each datapoint.

\section{Experiments}
\label{sec:experiments}

We implemented both minibatch approaches in PyTorch, and compared against the reference implementation from \citet{bovyExtremeDeconvolutionInferring2011}.
To evaluate each method, we used a random sample of rows from the Gaia DR2 source table~\cite{brownGaiaDataRelease2018}.
We selected the 5 primary astrometric features, along with the BP-RP colour and mean magnitude in the G-band.
In total there were 2 million rows.
Where data were missing, we set the field to zero and the noise variance to a large value.
We set the projection $R_i$ to the identity matrix for every sample.
This preliminary study uses only a small fraction of the full dataset size, but this allows us to fit the training data into memory, a requirement for use with the original implementation of extreme deconvolution.
We used a range of mixture component sizes $K$.
In practice we would want to select a value of $K$ by cross-validation.

The existing EM method ran on CPU, whilst the minibatch EM and SGD methods ran on GPU\@.
While the absolute times depend strongly on hardware and fine implementation details, they give a sense of the sort of times possible on current workstations, and the relative times across model sizes illustrate how the methods scale.
We used a validation set comprising $10\%$ of the rows when developing our experiments.
Final model performance was evaluated on a different held-out test set also comprising 10\% of the rows at the last stage, with no parameter selection or development done based on this set.
Details required for reproducibility are provided in Appendix~\ref{apx:repro}.

\begin{figure}
  \centering
  \includegraphics[width=\textwidth]{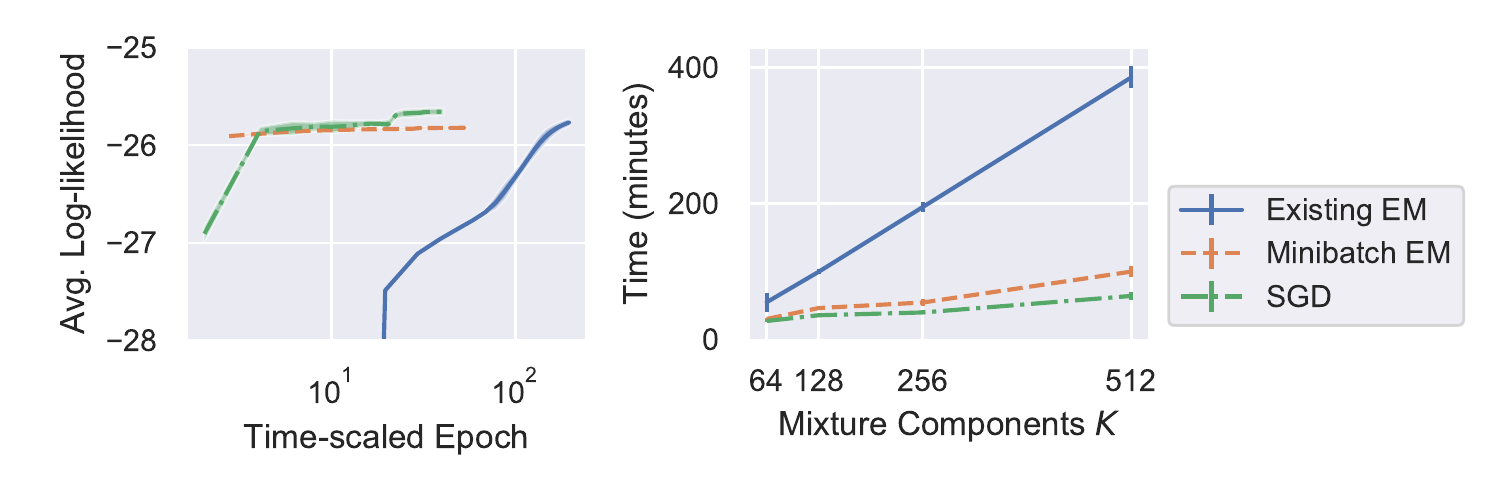}
  \caption{\textbf{Left}: Average training log-likelihood as a function of training on the Gaia subset for models with $K=256$. Epochs rescaled by average training time. Error bars not visible. \textbf{Right}: Training time as a function of mixture components $K$. Error bars indicate $\pm$ 2 standard deviations.}
  \label{fig:training}
\end{figure}

\begin{figure}
  \centering
  \includegraphics{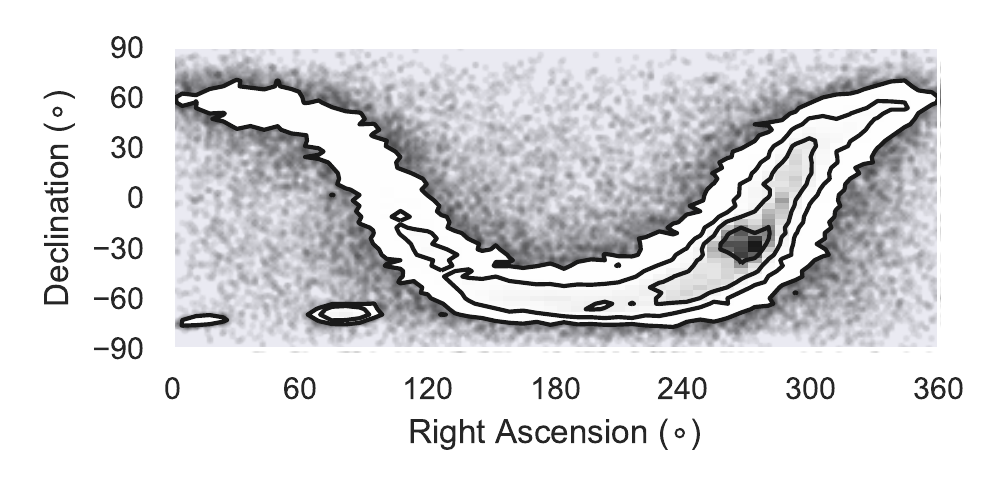}
  \caption{Density plot showing a 2-D projection of 100000 samples drawn from a model with $K=256$ and fitted with the minibatch EM method.
  The plot shows the estimated density of star positions on the sky, and has correctly recovered the structure of the Milky Way and the Magellanic Clouds.}
  \label{fig:projection}
\end{figure}

Table~\ref{results-table} reports the validation and test log-likelihoods for each method.
The values are similar, but not exactly comparable, as the effect of regularisation differs for each method.
Figure~\ref{fig:training} plots the training log-likelihood against time-rescaled epoch for $K=256$, and training time as function of mixture components $K$.
Figure~\ref{fig:projection} shows a 2-D projection from an example model with $K=256$ fitted with the minibatch-EM method.

\begin{table}{}
  \caption{Average validation log-likelihoods for the Gaia data subset for different numbers of mixture components $K$, with average test log-likelihood for the best value of $K$ by validation. Average over 10 runs with standard deviation.}
  \label{results-table}
  \centering
  \begin{tabular}{lcccc}
      \toprule
      Method     & K &  Validation     & Test\\
      \midrule
      Existing EM & 64 & $-26.10 \pm 0.03$ & - \\
      \citet{bovyExtremeDeconvolutionInferring2011} & 128 & $-25.96 \pm 0.04$ & - \\
       & 256 & $-25.76 \pm 0.02$ & - \\
       & 512 & $-25.67 \pm 0.01$ & $-25.66 \pm 0.01$ \\
      \midrule
      Minibatch EM & 64 & $-26.05 \pm 0.01$ & - \\
       & 128 & $-25.91 \pm 0.01$ & - \\
       & 256 & $-25.83 \pm 0.00$ & - \\
       & 512 & $-25.80 \pm 0.00$ & $-25.79 \pm 0.00$ \\
      \midrule
      SGD & 64 & $-25.89 \pm 0.02$ & - \\
       & 128 & $-25.77 \pm 0.02$ & - \\
       & 256 & $-25.67 \pm 0.02$ & - \\
       & 512 & $-25.59 \pm 0.02$ & $-25.57 \pm 0.02$ \\
      \bottomrule
  \end{tabular}
\end{table}

\section{Discussion}

Our results have shown that both of our proposed methods perform comparably to the existing method of fitting extreme deconvolution models, whilst converging faster.
The results also show that using GPU-based computation speeds up fitting considerably, allowing sublinear scaling of training time with mixture component size $K$. 

Further improvements to our approaches are possible.
The original paper presents a method of getting out of local maxima by splitting and merging clusters, with the split-merge criteria evaluated on the whole dataset.
It should be possible to replace the criteria with stochastic estimates, which would permit them to be used with both the SGD and minibatch EM methods.
Our approaches also add more free parameters to be selected, including learning rate and batch size.
This adds scope for hyperparameter optimisation to improve the log-likelihood.

In this preliminary study the SGD method provided the best log-likelihood values, was faster to train, and scaled better with component size $K$.
In addition, we found SGD to be more numerically stable than minibatch-EM during training.
Both minibatch methods will allow us to fit larger models going forwards.

\subsubsection*{Acknowledgments}
We are grateful to David W. Hogg for providing us with the problem and advising us on the Gaia data.
Our experiments made use of AstroPy~\citep{astropy:2013,astropy:2018}, corner.py~\citep{corner}, Matplotlib~\citep{Hunter:2007}, Numpy~\citep{numpy}, Pandas~\citep{mckinney-proc-scipy-2010}, PyTorch~\citep{paszke2017automatic} and Scikit-learn~\citep{scikit-learn}.
This work was supported in part by the EPSRC Centre for Doctoral Training in Data Science, funded by the UK Engineering and Physical Sciences Research Council (grant EP/L016427/1) and the University of Edinburgh.

\bibliography{references}

\begin{thebibliography}{19}
\providecommand{\natexlab}[1]{#1}
\providecommand{\url}[1]{\texttt{#1}}
\expandafter\ifx\csname urlstyle\endcsname\relax
  \providecommand{\doi}[1]{doi: #1}\else
  \providecommand{\doi}{doi: \begingroup \urlstyle{rm}\Url}\fi

\bibitem[Anderson et~al.(2018)Anderson, Hogg, Leistedt, {Price-Whelan}, and
  Bovy]{andersonImprovingGaiaParallax2018}
L.~Anderson, D.~W. Hogg, B.~Leistedt, A.~M. {Price-Whelan}, and J.~Bovy.
\newblock Improving {{Gaia Parallax Precision}} with a {{Data}}-driven
  {{Model}} of {{Stars}}.
\newblock \emph{The Astronomical Journal}, 156\penalty0 (4):\penalty0 145,
  Sept. 2018.

\bibitem[Bottou et~al.(2018)Bottou, Curtis, and Nocedal]{bottou2018}
L.~Bottou, F.~E. Curtis, and J.~Nocedal.
\newblock Optimization methods for large-scale machine learning.
\newblock \emph{SIAM Rev}, 60\penalty0 (2):\penalty0 223--311, 2018.

\bibitem[Bovy et~al.(2011)Bovy, Hogg, and
  Roweis]{bovyExtremeDeconvolutionInferring2011}
J.~Bovy, D.~W. Hogg, and S.~T. Roweis.
\newblock Extreme deconvolution: {{Inferring}} complete distribution functions
  from noisy, heterogeneous and incomplete observations.
\newblock \emph{The Annals of Applied Statistics}, 5\penalty0 (2B):\penalty0
  1657--1677, June 2011.

\bibitem[Capp{\'e} and Moulines(2009)]{cappeOnlineExpectationMaximization2009}
O.~Capp{\'e} and E.~Moulines.
\newblock On-line expectation\textendash{}maximization algorithm for latent
  data models.
\newblock \emph{Journal of the Royal Statistical Society: Series B (Statistical
  Methodology)}, 71\penalty0 (3):\penalty0 593--613, 2009.

\bibitem[Dempster et~al.(1977)Dempster, Laird, and
  Rubin]{dempsterMaximumLikelihoodIncomplete1977}
A.~P. Dempster, N.~M. Laird, and D.~B. Rubin.
\newblock Maximum {{Likelihood}} from {{Incomplete Data}} via the {{EM
  Algorithm}}.
\newblock \emph{Journal of the Royal Statistical Society. Series B
  (Methodological)}, 39\penalty0 (1):\penalty0 1--38, 1977.

\bibitem[Foreman-Mackey(2016)]{corner}
D.~Foreman-Mackey.
\newblock corner.py: Scatterplot matrices in python.
\newblock \emph{The Journal of Open Source Software}, 24, 2016.

\bibitem[Hunter(2007)]{Hunter:2007}
J.~D. Hunter.
\newblock Matplotlib: A 2d graphics environment.
\newblock \emph{Computing in Science \& Engineering}, 9\penalty0 (3):\penalty0
  90--95, 2007.

\bibitem[Kingma and Ba(2014)]{kingmaAdamMethodStochastic2014}
D.~P. Kingma and J.~Ba.
\newblock Adam: {{A Method}} for {{Stochastic Optimization}}.
\newblock \emph{3rd International Conference for Learning Representations},
  2014.

\bibitem[McKinney(2010)]{mckinney-proc-scipy-2010}
W.~McKinney.
\newblock Data structures for statistical computing in python.
\newblock In S.~van~der Walt and J.~Millman, editors, \emph{Proceedings of the
  9th Python in Science Conference}, pages 51 -- 56, 2010.

\bibitem[Oliphant(2006)]{numpy}
T.~Oliphant.
\newblock {NumPy}: A guide to {NumPy}.
\newblock USA: Trelgol Publishing, 2006.

\bibitem[Paszke et~al.(2017)Paszke, Gross, Chintala, Chanan, Yang, DeVito, Lin,
  Desmaison, Antiga, and Lerer]{paszke2017automatic}
A.~Paszke, S.~Gross, S.~Chintala, G.~Chanan, E.~Yang, Z.~DeVito, Z.~Lin,
  A.~Desmaison, L.~Antiga, and A.~Lerer.
\newblock Automatic differentiation in {PyTorch}.
\newblock In \emph{NIPS Autodiff Workshop}, 2017.

\bibitem[Pedregosa et~al.(2011)Pedregosa, Varoquaux, Gramfort, Michel, Thirion,
  Grisel, Blondel, Prettenhofer, Weiss, Dubourg, Vanderplas, Passos,
  Cournapeau, Brucher, Perrot, and Duchesnay]{scikit-learn}
F.~Pedregosa, G.~Varoquaux, A.~Gramfort, V.~Michel, B.~Thirion, O.~Grisel,
  M.~Blondel, P.~Prettenhofer, R.~Weiss, V.~Dubourg, J.~Vanderplas, A.~Passos,
  D.~Cournapeau, M.~Brucher, M.~Perrot, and E.~Duchesnay.
\newblock Scikit-learn: Machine learning in {P}ython.
\newblock \emph{Journal of Machine Learning Research}, 12:\penalty0 2825--2830,
  2011.

\bibitem[Perryman et~al.(1997)Perryman, Lindegren, Kovalevsky, H{\o}g, Bastian,
  Bernacca, Cr{\'e}z{\'e}, Donati, Grenon, Grewing, Van~Leeuwen, Van Der~Marel,
  Mignard, Murray, Le~Poole, Schrijver, Turon, Arenou, Froeschl{\'e}, and
  Petersen]{perrymanHipparcosCatalogue1997}
M.~A.~C. Perryman, L.~Lindegren, J.~Kovalevsky, E.~H{\o}g, U.~Bastian, P.~L.
  Bernacca, M.~Cr{\'e}z{\'e}, F.~Donati, M.~Grenon, M.~Grewing, F.~Van~Leeuwen,
  H.~Van Der~Marel, F.~Mignard, C.~A. Murray, R.~S. Le~Poole, H.~Schrijver,
  C.~Turon, F.~Arenou, M.~Froeschl{\'e}, and C.~S. Petersen.
\newblock The {{Hipparcos Catalogue}}.
\newblock \emph{Astronomy and Astrophysics}, 323\penalty0 (1):\penalty0 49--52,
  1997.

\bibitem[Sculley(2010)]{sculleyWebscaleKmeansClustering2010}
D.~Sculley.
\newblock Web-scale k-means clustering.
\newblock In \emph{Proceedings of the 19th International Conference on
  {{World}} Wide Web - {{WWW}} '10}, page 1177, {Raleigh, North Carolina, USA},
  2010. {ACM Press}.

\bibitem[{The Astropy Collaboration}(2013)]{astropy:2013}
{The Astropy Collaboration}.
\newblock {Astropy: A community Python package for astronomy}.
\newblock \emph{Astronomy \& Astrophysics}, 558:\penalty0 A33, Oct. 2013.

\bibitem[{The Astropy Collaboration}(2018)]{astropy:2018}
{The Astropy Collaboration}.
\newblock {The Astropy Project: Building an Open-science Project and Status of
  the v2.0 Core Package}.
\newblock \emph{The Astronomical Journal}, 156:\penalty0 123, Sept. 2018.

\bibitem[{The Gaia Collaboration}(2016)]{collaborationGaiaMission2016}
{The Gaia Collaboration}.
\newblock The {{Gaia}} mission.
\newblock \emph{Astronomy and Astrophysics}, 595:\penalty0 A1, Nov. 2016.

\bibitem[{The Gaia Collaboration}(2018)]{brownGaiaDataRelease2018}
{The Gaia Collaboration}.
\newblock Gaia {{Data Release}} 2. {{Summary}} of the contents and survey
  properties.
\newblock \emph{Astronomy \& Astrophysics}, 616:\penalty0 A1, Aug. 2018.

\bibitem[Williams(1996)]{williams1996}
P.~M. Williams.
\newblock Using neural networks to model conditional multivariate densities.
\newblock \emph{Neural Computation}, 8\penalty0 (4):\penalty0 843--854, 1996.

\end{thebibliography}

\appendix

\section{Stable Covariance Update}
\label{apx:variance-rewrite}

In Section~\ref{sec:minibatch-em} describing our minibatch-EM method, we noted that the M-step update for the variance of each component $V_j$ as presented in Equation~\ref{eqn:mstep} would be prone to catastrophic cancellation as a result of taking a small difference between large values using single-precision floats.
Here we present an alternative update for $V_j$ that is less prone to numerical instability, and show that it is equivalent to Equation~\ref{eqn:mstep}.
For clarity we drop the component indicator $j$ from the parameters, and add indicators $t$ and $t-1$ to distinguish between current and previous estimates of parameters.

First, we define an adjustment operation,
\begin{align}
  \text{adjust}(V, s, \bc, \bd) &= sV + \frac{1}{2}(\bc - \bd)(\bc + \bd)^T + \frac{1}{2}(\bc + \bd)(\bc - \bd)^T  \label{eq:recentre1} \\
  &= s(V + \bc\bc^T) - \bd\bd^T,\label{eq:recentre2}
\end{align}
which can be thought of as recentering a scaled variance around a new mean.
Equation~\ref{eq:recentre1} is how we actually compute the adjustment, to minimise taking small differences between large values, whilst Equation~\ref{eq:recentre2} shows the identity we are interested in.

In the M-step at iteration $t$ of our minibatch EM approach, we compute estimates of $\hat{q}_t$, $\alpha_t$ and $\bm_t$ as before using Equations~\ref{eq:sums} and~\ref{eqn:mstep}.
We also compute minibatch-specific parameters using exact sums over the minibatch:
\begin{equation}
q_b = \sum_i^M r_i, \quad
\bm_b = \frac{\sum_i^M r_i \bx_i}{q_b},\quad V_b = \frac{\sum_i^M r_i[(\bx_i - \bb_i)(\bx_i - \bb_i)^T  + B_i]}{q_b}
\end{equation}{}
We then compute our new estimate of the variance $V_t$ as a function of the previous estimates $\{\hat{q}_{t-1}, \bm_{t-1}, V_{t-1} \}$, the minibatch values $\{q_b, \bm_b, V_b\}$, and the current estimates $\{\hat{q}_t,\bm_t \}$:
\begin{align}
V_{t} &= (1 - \lambda)\,\text{adjust}(V_{t-1}, \frac{\hat{q}_{t-1}}{\hat{q}_t}, \bm_{t-1}, \bm_t) + \lambda\, \text{adjust}(V_{b}, \frac{q_b}{\hat{q}_t}, \bm_{b}, \bm_{t}) \label{eq:update-computed}\\
&= (1 - \lambda) \left[\frac{\hat{q}_{t-1}}{\hat{q}_t} \left(V_{t-1} + \bm_{t-1}\bm_{t-1}^T \right) -\bm_t \bm_t^T \right] + \lambda\,\left[\frac{q_b}{\hat{q}_t} \left(V_{b} + \bm_{b}\bm_{b}^T \right) -\bm_t \bm_t^T \right] \\
&= (1 - \lambda)\,\left[\frac{\hat{S}_{t-1}}{\hat{q}_t} -\bm_t \bm_t^T \right] + \lambda\, \left[\frac{S_t}{\hat{q}_t} -\bm_t \bm_t^T \right] \\
&= \frac{(1- \lambda)\hat{S}_{t-1} + \lambda S_t}{\hat{q}_t} - \bm_t\bm_t^T \\
&= \frac{\hat{S}_t}{\hat{q}_t} - \bm_t\bm_t^T \label{eq:update-equiv}
\end{align}
Again, Equation~\ref{eq:update-computed} is how we actually compute the update to minimise numerical errors, whilst Equation~\ref{eq:update-equiv} shows that the update is equivalent to the covariance update defined in Equation~\ref{eqn:mstep}.
Whilst we found this update worked better in practice than a direct implementation, numerical instability is still possible if the standard deviations of the components are small enough relative to the means, and further work is needed to determine if a more stable update can be performed.

\section{Experiment Details}
\label{apx:repro}

Here we provide specific details of our experiments for reproducibility.
Code used to run the experiments is available at \url{https://github.com/bayesiains/scalable_xd}.

\subsection{Dataset}

From the Gaia DR2 source table we selected the columns \texttt{RA}, \texttt{DEC}, \texttt{PARALLAX}, \texttt{PMRA}, \texttt{PMDA}, \texttt{BP\_RP} and \texttt{PHOT\_G\_MEAN\_MAG} to assemble the observed dataset $\{\bx_i\}_{i=0}^N$~\cite{brownGaiaDataRelease2018}.
Random subsampling was done by selecting rows with the value of the \texttt{RANDOM\_INDEX} column less than $\numprint{2000000}$.
Noise covariance matrices $S_i$ were assembled using the corresponding error and correlation columns for each variable.
Where a column of a row was marked as missing, the corresponding element of $\bx_i$ was set to zero, the corresponding diagonal element of $S_i$ was set to $10^{12}$, and the corresponding off-diagonal elements set to zero.
For columns which do not have associated noise, the corresponding diagonal elements of $S_i$ were set to a value of $10^{-2}$, and corresponding off-diagonal elements to zero.

\subsection{Initialisation}
For each method, initialisation of the means and weights was done using the estimated counts and centroids after 10 epochs of minibatch k-means clustering~\cite{sculleyWebscaleKmeansClustering2010}.
Covariances $V_j$ were set to the identity matrix.

\subsection{Training}

All methods were trained for a total of twenty epochs.
Both minibatch methods used a batch size of $500$.
For the minibatch-EM method, the step size $\lambda$ was started at $10^{-2}$ and reduced by a factor of two after ten epochs.
For the SGD method, we used the Adam optimiser with a learning of $10^{-2}$ for the first ten epochs, reducing by a factor of ten for the last ten epochs, and all other parameters set to the suggested defaults~\cite{kingmaAdamMethodStochastic2014}.

For numerical stability, a very small amount of regularisation was applied to the covariances $V_j$.
Using the original implementation, we set the regularisation constant $w = 10^{-3}$.
For the minibatch-EM method, we added diagonal matrix $wI$ directly to each covariance matrix after updating them.
For the SGD method, we added a penalty term $\sum_j \frac{w}{\mathrm{Trace}[V_j]}$ to the loss function.
As noted in Section~\ref{sec:experiments}, the effect of $w$ is not comparable across methods.
If we were using larger values of $w$ to prevent overfitting rather than just avoiding numerical instability, $w$ would be tuned specifically for each method.

\end{document}